\documentclass[letterpaper, 10 pt, journal, twoside]{ieeetran}

\IEEEoverridecommandlockouts
\let\labelindent\relax

\usepackage{graphics}
\usepackage{graphicx}
\usepackage{epsfig}
\usepackage{times}
\usepackage{amsmath}
\usepackage{amssymb}
\usepackage[ruled,vlined,linesnumbered]{algorithm2e}
\usepackage{algpseudocode}
\usepackage{svg}
\usepackage{multirow}
\usepackage{tikz}
\usepackage{enumitem}
\usepackage{booktabs}
\usepackage{url}
\usepackage{hyperref}

\title{MANER: Multi-agent Neural Rearrangement Planning of Objects in Cluttered Environments}
\author{Vivek Gupta, Praphpreet Dhir, Jeegn Dani, and Ahmed H. Qureshi
\thanks{The authors are with the Purdue University, West Lafayette, IN 47907, USA
        {\tt\footnotesize \{gupta690, pdhir, jdani, ahqureshi\}@purdue.edu}}%
\thanks{V.Gupta and P.Dhir did the project during their stay at Purdue University.}
}%Use only for final RAL version

\def\R{{\mathbb{R}}} % real numbers
 
\begin{document}

\maketitle
%\thispagestyle{empty}
%\pagestyle{empty}

%%%%%%%%%%%%%%%%%%%%%%%%%%%%%%%%%%%%%%%%%%%%%%%%%%%%%%%%%%%%%%%%%%%%%%%%%%%%%%%%
\begin{abstract}
%Object rearrangement is a fundamental problem in robotics with various practical applications in warehouse automation, robotic kitchens, and precision manufacturing. While existing research has primarily focused on single-agent solutions, real-world scenarios often require multiple robots to work together on rearrangement tasks. In this paper, we propose a comprehensive framework for multi-agent object rearrangement planning, addressing the challenges of task sequencing and path planning in complex environments. We propose an integrated task and motion planning approach that considers the environment, robot kinematics, and task reachability. Our experiments on a diverse range of environments demonstrate the effectiveness and robustness of the proposed framework. Results indicate improved performance in terms of traversal time and success rate compared to baseline approaches.
Object rearrangement is a fundamental problem in robotics with various practical applications ranging from managing warehouses to cleaning and organizing home kitchens. While existing research has primarily focused on single-agent solutions, real-world scenarios often require multiple robots to work together on rearrangement tasks. This paper proposes a comprehensive learning-based framework for multi-agent object rearrangement planning, addressing the challenges of task sequencing and path planning in complex environments. The proposed method iteratively selects objects, determines their relocation regions, and pairs them with available robots under kinematic feasibility and task reachability for execution to achieve the target arrangement. Our experiments on a diverse range of simulated and real-world environments demonstrate the effectiveness and robustness of the proposed framework. Furthermore, results indicate improved performance in terms of traversal time and success rate compared to baseline approaches. The videos and supplementary material are available at %\url{https://sites.google.com/view/maner-framework/home} 
\url{https://sites.google.com/view/maner-supplementary}
\end{abstract}


%%%%%%%%%%%%%%%%%%%%%%%%%%%%%%%%%%%%%%%%%%%%%%%%%%%%%%%%%%%%%%%%%%%%%%%%%%%%%%%%
\section{INTRODUCTION}
\IEEEPARstart{O}{bject} rearrangement is a fundamental problem in the field of robotics, with diverse applications in warehouse automation, robotic kitchens, and precision manufacturing. The ability to rearrange objects in a desired configuration is crucial for improving efficiency, organization, and productivity in these domains. The problem involves three stages: task planning, path planning, and execution. During the first phase, the robot decomposes the task into sub-tasks while considering the environment state, robot kinematic reachability, and task feasibility. The remaining phases find the robot's motion path and control actions to achieve the given subtasks in the environment. While most of the existing work addresses the rearrangement planning with a robot arm in tabletop environments, the solutions via multi-agent mobile manipulation systems remain largely unexplored.

In multi-agent systems, task planning plays a crucial role in achieving a desired global objective by assigning optimal sequences of sub-goals to each agent. It involves solving a combinatorial optimization problem, which determines the most efficient order of tasks for agents to accomplish. This stage is a vital precursor to path planning and execution. Existing research has proposed solvers to address some of these planning problems \cite{hartmann2022long} \cite{wu2021spatial}. However, these approaches do not consider noisy observations, rely on simplified environments, and assume objects have trackable markers and readily reachable goals, i.e., monotone scenarios. In contrast, the practical scenarios often lack objects with trackable markers and contain non-monotone cases, where objects require multiple relocations before making it to the goal, adding complexity to the problem of rearrangement planning and making vision-based techniques necessary.

%Object rearrangement is a fundamental problem in the field of robotics with diverse applications in warehouse automation, robotic kitchens, and precision manufacturing. The ability to rearrange objects in a desired configuration is crucial for improving efficiency, organization, and productivity in these domains. The problem involves three stages: task planning, path planning, and execution. During the first phase, the robot decomposes the task into sub-tasks and aims to find a sequence that approaches the desired configuration. This process considers factors such as the state of environment, robot kinematics, and task feasibility. Non-monotone cases, where objects require multiple transfers, add significant complexity to the problem. While existing planning approaches address single manipulator systems, multi-agent systems remain largely unexplored.

%In multi-agent systems, task planning plays a crucial role in achieving a desired global objective by assigning optimal sequences of sub-goals to each agent. It involves solving a combinatorial optimization problem, which determines the most efficient order of tasks for agents to accomplish. This stage is a vital precursor to path planning and execution. Existing research has proposed algorithmic solvers and learning-based approaches to address some of these planning problems \cite{hartmann2022long} \cite{wu2021spatial}. However, these approaches often assume simplified environments with readily reachable goals. Moreover, object rearrangement planning has received limited attention in the context of multi-agent systems.

In this paper, we present \textbf{M}ulti-\textbf{A}gent \textbf{Ne}ural \textbf{R}earrangement (MANER) framework for long-horizon multi-agent object rearrangement planning from visual observations in cluttered environments. The framework analyzes the current and target state of the environment to determine potential pickup objects for each agent. Additionally, it proposes feasible drop-off locations for each agent-object pair. Subsequently, a path planning framework evaluates these pick-place propositions to select the best pick-place sequence, enabling efficient scene arrangement by multiple agents. The key contributions of this paper are:
\begin{itemize}[leftmargin=*]
\item a pick and place module to compute the feasible sub-task sequences for each agent, considering robot traversal time, reachability, and collision avoidance.
\item a scalable rearrangement planning framework that accommodates the varying number of objects and agents in the scene and tackles both monotone and non-monotone cases.
\item a training and testing strategy that enables one-shot sim2real transfer of our neural policies to solve multi-agent rearrangement planning problems in real-world scenarios, relying only on the bird's eye view of the environment.
\end{itemize}

To the best of our knowledge, MANER is the first learning-based approach that specifically tackles the task of multi-agent object rearrangement planning, accounting for complex rearrangement scenarios where objects may require multiple repositioning steps. Our experimental results demonstrate the effectiveness and superior performance of MANER compared to classical baseline methods and its ability to generalize to real-world scenarios.

% The structure of this paper is as follows. In Section II, we review the existing literature on object rearrangement planning and multi-agent systems. Section III presents the problem statement and introduces the MANER framework, with a focus on the pick, place, and path planning module. In Section IV, we provide implementation details, including data generation and training procedures. Section V presents the results and analysis of our approach. Finally, in Section VI, we conclude the paper by highlighting the significance of MANER in advancing multi-agent object rearrangement planning in robotics and outlining potential future research directions.

\section{RELATED WORK}
\subsection{Rearrangement Planning}
Rearrangement planning has garnered significant attention from researchers due to its applicability in various domains \cite{batra2020rearrangement}. Tree search- and learning-based approaches have been widely explored to solve this problem. For instance, methods \cite{labbe2020monte, krontiris2015dealing} introduce a tree search-based strategy to mitigate the combinatorial complexity in decision-making for rearrangement planning. In a similar vein, Ren et al. \cite{ren2022rearrangement} proposed a rearrangement-based manipulation approach using kinodynamic planning with dynamic horizons. While these tree search-based methods have demonstrated effectiveness to some extent, they exhibit poor asymptotic performance as the environment becomes cluttered with more objects. 

In contrast, the learning-based approaches \cite{zeng2021transporter} have recently emerged as a scalable tool for addressing the rearrangement planning problem. Qureshi et al. \cite{qureshi2021nerp} proposed a neural approach that estimates the optimal pick-and-place actions for each task in the rearrangement problem. Their work incorporates reactive object placement strategies and exhibits promising results. Wu et al. \cite{wu2022transporters} introduced a visual foresight model that leverages deep learning to generalize to real-world environments, enabling zero-shot rearrangement planning. However, all the abovementioned methods, including classical and learning-based, have predominantly focused on single-agent manipulation systems with minimal kinematic reachability challenges, overlooking the complexities of multi-agent mobile manipulation systems.

\subsection{Multi-Agent Systems}
In the field of multi-agent systems, the concept of coordinating multiple agents to achieve distributive and cooperative tasks has a longstanding history, tracing back to the 1980s \cite{MultiMobileRobotSystems}. Since then, several approaches have been introduced, broadly falling into search, optimization, and learning-based methods. The search-based methods mainly focus on strategies to reduce the search space for multi-robot task allocation \cite{levihn2012multi}. Recently, Motes et al. \cite{motes2022hypergraph} introduced a hypergraph representation for multi-robot task and motion planning, demonstrating its effectiveness in generic rearrangement tasks. Gao et al. \cite{gao2022toward} proposed a dependency graph-guided heuristic search procedure tailored to non-monotone rearrangement tasks, albeit considering only two manipulators. Still, despite advancements, the search-based approaches lack scalability to a large number of objects or robots due to their computational cost. 

The optimization-based approaches solve the constraint optimization problem for multi-robot task allocation and scheduling and have surfaced as a promising tool for solving various assembly tasks \cite{hartmann2022long,chen2022cooperative}. However, the existing work considers monotone cases with readily reachable goals, assumes complete information about the environmental states, and is yet to be explored in real-world settings with raw visual observations. In the realm of learning-based methods, the problem of rearrangement planning with multi-agent mobile manipulation systems remains relatively unexplored, except for recent work \cite{wu2021spatial}. This approach also considers only monotone cases where the task is to transport objects to a fixed target. Their intention-based representation and deep reinforcement learning framework showed promise in coordinating multiple mobile robots to achieve such tasks. However, their approach assumes the known location of the robots and objects, i.e., it does not operate under noisy visual observations. 

In contrast to the abovementioned approaches, our method solves both monotone and non-monotone rearrangement planning problems. In addition, our framework operates directly from raw visual observation and generalizes to real-world settings via direct sim2real transfer. Besides, the tasks considered in this work are significantly more challenging than previous works, requiring planners to consider goal reachability as goals may initially be occupied, feasible space due to static and dynamic obstacles, and robot kinematics.
% Divide this into two sections: Problem Statement and Proposed method
\section{Proposed Method: MANER}
Given a bounded workspace $W \subset  SE(2)$, containing $m \in \mathbb{N}$ agents and $n \in \mathbb{N}$ movable objects, and a set of static obstacles, our goal is to find a motion plan for each agent that rearranges the environment to achieve the target configuration. %We assume that there are fewer agents than movable objects in the workspace, i.e., $m < n$.

We define our multi-agent rearrangement planning problem as follows: The target scene of the environment is denoted by $X_T$, and the current scene at time $t$ is denoted by $X_t$. The states of objects in the current and target scene images are denoted as $o^t_{\{n\}}$ and $o^T_{\{n\}}$ respectively. Here, $o\in o_{\{n\}}$ represents the object state comprising position and instance label. The multi-agent robot states at time $t$ are denoted as $r^t_{\{m\}}$, and $r^t \in r^t_{\{m\}}$ is the three-dimensional pose of an agent at time $t$. Let $l_{\{b\}}$ comprise all the free regions in the current scene $X_t$. The multi-agent object rearrangement planning problem involves pairing each available agent with an object from the scene and allocating a free region for its relocation. Each agent picks its paired object and relocates it to the assigned region. This process is repeated until the target arrangement in $X_T$ is achieved. Therefore, at each time step $t$, the MANER policy, defined as $\pi$, takes as input the current and target scene and outputs the action $\tau^t=\{(r^t_0, o^t_0, l^t_0),\cdots, (r^t_i, o^t_j, l^t_k) \}$, where $i\in [0,m]$, $j\in [0,n]$, $k\in [0,b]$, and each element $(r^t, o^t, l^t)$ is best pairing of a robot at state $r^t$ with pick-up object $o^t$, and its placement location $l^t$. In other words,
\begin{equation*}
\{(r^t_0, o^t_0, l^t_0),\cdots, (r^t_i, o^t_j, l^t_k) \} \longleftarrow \pi(X_t, X_T)
\end{equation*}

\begin{figure*}
\vspace{0.02in}
  \includegraphics[width=\textwidth, trim={0 0.4cm 0 0}, clip]{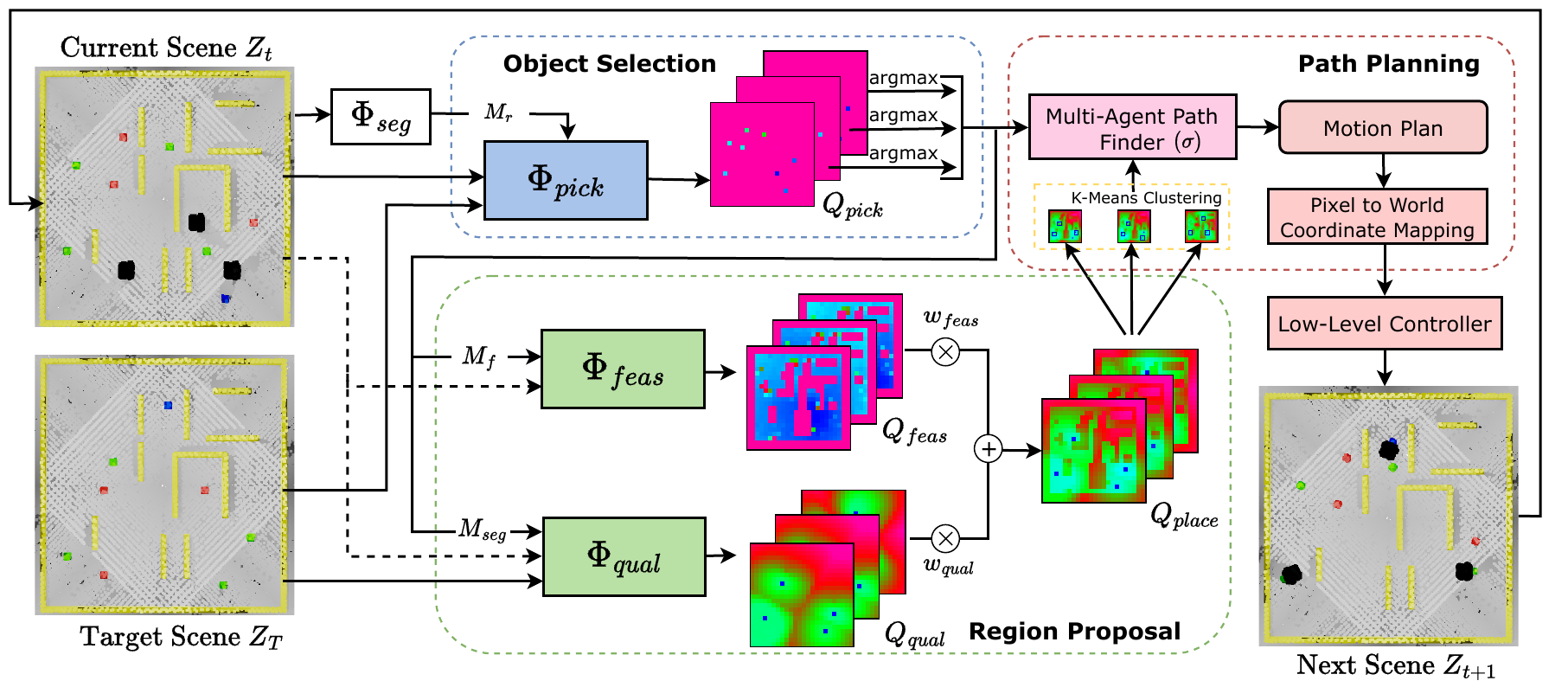}
  \vspace{-0.7cm}
  \caption{MANER architecture: Initially, bird's-eye view images of the current and target scenes are captured and processed. These images $Z_t$ and $Z_T$ are then passed through the Object Selection Module, denoted as $\Phi_{pick}$, which generates a heatmap called $Q_{pick}$. This output is further processed by the region proposal module, which consists of two transformer networks, $\Phi_{feas}$ and $\Phi_{qual}$ and works together to produce a final heatmap named $Q_{place}$ identifying suitable regions for object placement. The proposed regions are clustered using the $k$-means algorithm, and then fed into a Multi-Agent Path Planner ($\sigma$), which generates a motion plan. Subsequently, the visual coordinates are converted to world coordinates and the motion plan is executed to reach the next state $Z_{t+1}$. Throughout the planning and execution phases, a low-level controller is utilized to guide the agents and achieve the desired object rearrangement. Note that the darker pixels in the heatmap indicate higher confidence and lighter pixels indicate lower confidence.}
  
  \label{fig:maner}
  \vspace{-0.6cm}
\end{figure*}

We introduce a function $\sigma$ that represents an underlying multi-agent motion planner. The function $\sigma$ takes the given pick-and-place action sequence $\tau^t$ and determines if there exist collision-free path solutions for each robot $r^t$ to reach its target object $o^t$ and then the relocation region $l^t$. The function $\sigma$ returns $\emptyset$ if any of the required paths are not found in the given time limit. Let $E$ be the environment function that executes the given path $\sigma(\tau^t)$ on $X_t$ and returns the next environment scene $X_{t+1}$. Furthermore, we assume all robots execute their assigned subtasks synchronously and, after completion, wait for the subsequent subtask assignments. Finally, let $F$ indicate the total traversal time for all robots in executing their assigned tasks in $\tau^t$.

Our objective is to find an near-optimal function $\pi$ that solves the multi-agent rearrangement problem. We formulate our objective as follows:
\begin{equation*}
\begin{aligned}
\min_{\tau^t \sim \pi} & \sum_{t=0}^{T-1}{F(\tau^{t})}, \;\; \text{subject to}\\
&X_T=E(\sigma(\tau^{T-1}),X_{T-1})\: \& \: \sigma(\tau^t)\neq \emptyset\: \forall t\in [0,T-1]\\
  %&\sigma(\tau^0, \cdots, \tau^{T-1})\neq \phi    \\
\end{aligned}
\end{equation*}

We aim to generate an action trajectory $\{\tau^0, \cdots, \tau^{T-1}\}$ through the function $\pi$ that minimizes the overall multi-robot travel time given by $F$. Additionally, the output trajectory should lead to feasible multi-robot motion planning, i.e., $\sigma(\tau^t)\neq \emptyset \: \forall t \in [0,T-1]$, and result in achieving the target scene arrangement $X_T$. The following sections provide a detailed description of MANER, while Figure. \ref{fig:maner} and Algorithm \ref{alg:maner-main} provides an overview of the overall framework.

\subsection{Scene Preprocessing}
% Describe scene segmentation and correspondence matching.
To capture the entire state $X_t$ of the environment in our environments, we utilize RGBD camera(s). In case of multiple cameras, the captured images are converted to a single point cloud using the intrinsic property $K$, extrinsic property $R_t$, and depth information provided by those cameras. We take the bird's eye view image of the reconstructed point cloud to obtain the processed environment state $Z_t$. Next, we utilize an instance segmentation model $\Phi_{seg}$ \cite{wu2019detectron2} \cite{he2017mask} to extract relevant object states $o^t_{\{n\}}$ from the RGB image $Z_t$.

\subsection{Neural Models} \label{sec:neural}
% Change the name of the subsection maybe?
MANER consists of two neural network modules that enable efficient performance of the object rearrangement task even in cluttered environments. These modules generate the action sequence $\tau^t$ that specifies the pick-and-place actions for each agent. These components are described as follows.

\subsubsection{Object Selection Module}
The object selection module is responsible for determining the most suitable object for each agent to relocate. This module takes into account various factors, including the current scene arrangement, target scene arrangement, object states, and the positions of the agents, to identify objects that meet the criteria of being in close proximity to the agents and feasible to be picked.

The module $\Phi_{pick}$ takes as input a stacked map $M_{pick} \in \R^{H \times W \times 7}$ which includes the top-down observation of the current scene, denoted as $Z_t \in \R^{H \times W \times 3}$, the target scene $Z_T \in \R^{H \times W \times 3}$, and a map $M_r \in \R^{H \times W \times 1}$ encoding the current state of each agent $r$ in the arena. Using this information, it  generates a heatmap $Q_{pick} \in \R^{H_l \times W_l \times m}$, where $H_l = H / p$ and $W_l = W / p$. Each pixel $(u, v, r)$ in $Q_{pick}$ corresponds to a squared patch of size $p \times p$ in the current environment image, representing the confidence level of agent $r$ to successfully transition and pick an object from that location (see Figure \ref{fig:maner}). The agents are assigned a unique location based on the highest values in $Q_{pick}$, reflecting the most promising pick-up spots in the scene. To determine the specific object that each agent is prepared to pick, the module utilizes the object states $o^t_{\{n\}}$ extracted from $Z_t$ and the promising pick-up spots for each agent. The final output is a selection of agent-object pairs, i.e., $\tau_{pick}=\{(r^t_0, o^t_0),\cdots, (r^t_i, o^t_j)\}$, where $i\in [0,m]$ and $j\in [0,n]$. 

\subsubsection{Region Proposal Module}
This component consists of two interconnected networks: the Feasibility Network and the Quality Network. These networks work together to identify optimal regions for each agent-object pair, taking into account both the feasibility of reaching the region and the importance of the region for object relocation in terms of goal reachability. The networks are discussed below.

\paragraph {Feasibility Network}
Once an object is selected, the next step involves transferring it to an optimal location that the agent can reach in the shortest possible time. To achieve this, the feasibility network plays a crucial role. The feasibility network $\Phi_{feas}$ takes a stacked map $M_{feas} \in \mathbb{R}^{H \times W \times 2}$ as input. This map consists of a binary image of the current scene, denoted as $B_t \in \mathbb{R}^{H \times W \times 1}$, where free locations are represented by 0, and static obstacles and objects in the arena are represented by 1. Additionally, the map includes $M_f \in \mathbb{R}^{H \times W \times 1}$, which specifies the position of the object to be picked by the current agent and that of the other agents as provided by the Object Selection Module. The location of the current object is denoted by +1, while that of other picked objects is represented by -1.

The feasibility network then produces a heatmap denoted as $Q_{feas} \in \mathbb{R}^{H_l \times W_l \times m}$. This heatmap indicates the feasibility or reachability of different locations on the map for each agent. The feasibility score takes into account factors such as traversal time and location sparsity, aiming to identify regions that can be efficiently accessed while avoiding obstacles and other agents. If an accessible region is densely occupied, the feasibility score is lower compared to regions with fewer nearby objects. This ensures that the agent can transfer the object efficiently to the selected location while navigating around obstacles and other agents.

\paragraph{Quality Network}
When determining the best regions for transferring an object, we need to consider not only the reachability but also the proximity to possible final locations. For instance, if an object cannot be directly transferred to a final destination and needs to be moved to an intermediate location, regions closer to the final destination should be better suited for it than those farther away. To accomplish this, the quality network $\Phi_{qual}$ is employed. The input to this network consists of stacked map $M_{qual} \in \R^{H \times W \times 9}$ consisting of top-down observation of the current scene, denoted as $Z_t \in \R^{H \times W \times 3}$ and target scene $Z_T \in \R^{H \times W \times 3}$, and a segmentation mask $M_{seg} \in \R^{H \times W \times 3}$ of the current scene from the object selection module $Q_{pick}$. It generates a heatmap $Q_{qual} \in \R^{H_l \times W_l \times m}$, indicating the quality score of each patch for object transfer. This score accounts for the proximity to the target location, aiming to minimize unnecessary movements during the process.

To determine the best transfer regions for each agent-object pair, we employ a weighted sum $Q_{place}$ that combines the outputs from the Feasibility Network and the Quality Network. $Q_{place}$ is computed according to Line \ref{alg:maner-main:place} in Algorithm \ref{alg:maner-main}. The weights $w_{f} \in (0, 1)$ and $w_{q} \in (0, 1)$ are used to control the relative importance of feasibility and quality, respectively, and can be adjusted based on the specific task requirements. For our experiments, we set $w_{f}$ to 0.2 and $w_{q}$ to 0.8 as a balanced configuration.

Instead of selecting a single region for each agent-object pair, the region proposal module applies techniques such as $k$-means clustering to the output heatmap $Q_{place}$, generating a set of $k$ potential transfer regions denoted as $\tau^r_{place} \in l_{{b}}$ (Line \ref{alg:maner-main:taur}). $\alpha$ is a hyperparameter in the range of $(0, 1)$, used to filter out regions with insignificant values (Line \ref{alg:maner-main:alpha}). These proposed transfer regions represent possible intermediate or final locations where the agent can successfully transfer the object. In our experiments, we propose three regions for each agent-object pair, i.e., $k = 3$.

\begin{algorithm}[h]
\caption{MANER ($X_0$, $X_T$)}
\label{alg:maner-main}

$Z_T \gets \Call{\small ScenePreprocessing}{X_T}$

\For{$t = 0, \ldots, T-1$} {
   $Z_t \gets \Call{\small ScenePreprocessing}{X_t}$

  $B_t \gets \Call{\small BinaryImage}{Z_t}$

  $o^t_{\{n\}}, r^t_{\{m\}} \gets \Call{\small InstanceSegmentation}{Z_t}$
   
  $\tau_{pick} \gets [], \tau_{place} \gets []$
    
  % pathlen $\gets \infty$, robjects $\gets 0$

  % Initialize Motion Plan $\mathcal{A}$ to $\phi$

  \For{each agent $r \in r^t_{\{m\}}$} {
    $M_r \gets \Call{\small AgentEncode}{r, r^t_{\{m\}} \symbol{92} r}$

    $Q_{pick} \gets \Phi_{pick}(Z_t, Z_T, M_r)$

    $o_{pick}(r) \gets \arg \max_{(u, v)} {Q_{pick}}(u, v, r)$

    $\tau_{pick} \gets \tau_{pick} \cup (r, o_{pick}(r))$\

  }
  \For{each $(r, o_{pick}) \in \tau_{pick}$}{
    $M_f \gets \Call{\small ObjectEncode}{o_{pick}, \tau_{pick}}$

    $M_{seg} \gets \Call{\small SegmentationMask}{Z_t, o_{pick}}$

    $Q_{feas} \gets \Phi_{feas}(B_t, M_f)$

    $Q_{qual} \gets \Phi_{qual}(Z_t, Z_T, M_{seg})$

    $Q_{place} \gets {w_{f}*Q_{feas} + w_{q}*Q_{qual}}$ \label{alg:maner-main:place}

    $Q^{max}_{place}(r) \gets \arg \max_{(u, v)} {Q_{place}(u, v, r)}$

    $S \gets \phi$

    \For{each $(u, v)$} {
        \If{$Q_{place}(u, v, r) > \alpha Q^{max}_{place}(r)$} { \label{alg:maner-main:alpha} 
            append $(u, v)$ to $S$
        }
    }

    $\tau^r_{place} \gets \Call{\small K\_Means\_Cluster}{S}$ \label{alg:maner-main:taur} 

    $\tau_{place} \gets \tau_{place} \cup \tau^r_{place}$
  }
 
  % \For{each $\tau'_{place}\in \tau_{place}$} {
    
  %   $\mathcal{A'}$, len, objects = $\Call{\small \sigma}{\tau_{pick}, \tau'_{place}}$

  %   $success = \text{objects} > \text{robjects } \& \text{ len} < \text{pathlen}$

  %   \If{success} {
  %       $\mathcal{A}$ = $\mathcal{A'}$\;
  %       pathlen =  len, robjects = objects\;
  %   }
  % }

    $\tau \gets \Call{\small PathPlanning}{\tau_{pick}, \tau_{place}}$

    $X_{t+1} \gets E(\sigma(\tau), X_t)$}

\end{algorithm}

\subsection{Path Planning}
Once the Object Selection module and Region Proposal module have determined the objects to be picked and the potential transfer regions, the next step is to plan the paths for each agent to execute the rearrangement actions. For our experiments, we employed multi-agent path planning algorithm called AA-SIPP(m) (Anytime Any-angle Search with Safe Interval Path Planning for Multi-Agent Systems) \cite{yakovlev2017any}. This algorithm is designed for multi-agent path planning in shared 2D workspaces and can find collision-free paths for multiple agents operating simultaneously.

%TODO: fix the tuple% 
In this path planning phase, we utilize the multi-agent path planner $\sigma$ to find the near-optimal motion plan for $\tau$ that is free from collisions if such a plan exists. Each element in $\tau$ is of the form $(r, o, l)$, where $r$ represents the location of an agent, $o$ denotes the assigned object, and $l$ indicates the selected transfer region for the agent-object pair. Given that we have defined multiple possible regions $\tau^r_{place}$ for each agent-object pair, our goal is to select the transfer regions $\tau^{max}_{place} \in l_{b}$ for each agent that results in the shortest collision-free path while maximizing the number of objects successfully relocated to their desired target locations. This process ensures that the action sequences generated by the Object Selection and Region Proposal modules can be executed effectively, enabling the agents to navigate the environment freely without any collisions. The low-level controller then facilitates the execution of the resulting motion plan after pixel-to-world coordinate conversion using the camera's intrinsic, extrinsic, and depth information, leading to the next state.

By incorporating the key components described above, MANER offers a comprehensive solution for multi-agent object rearrangement planning. It effectively addresses the objective of minimizing the overall multi-agent traversal time while sequencing each agent's tasks, ensuring feasible motion planning, and achieving the target scene arrangement.

% \subsection{Execution pipeline}

\section{Implementation Details}
\subsection{Environment Setup} \label{section:env-setup}
\begin{figure}[ht]
\centering
\includegraphics[trim={0 0.2cm 0 0}, scale=1.03]{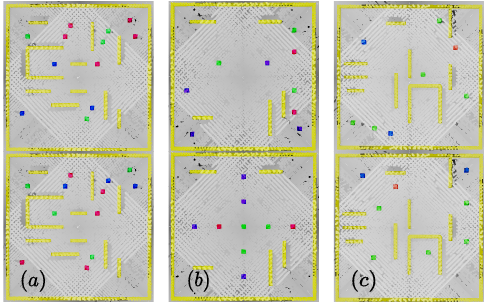}
\caption{Start (top) and target (bottom) configuration for different types of rearrangement tasks. These include (a) shuffling objects, (b) sorting objects, and (c) random placement of objects.}
\label{fig:tasks}
\vspace{-0.5cm}
\end{figure}

In both simulation and real environment setups, we employ Turtlebot3 Robots, which are built on the ROS standard platform, for object manipulation. To capture the state of the environment at every time step, we utilize RGBD camera(s). For simulation experiments, we use four such synthetic cameras, while for real experiments, we use Intel Realsense D455 camera. Each environment is composed of varying numbers of objects, agents, and static obstacles, each with diverse shapes and sizes. The locations of these objects, agents and obstacles are randomized in each experiment. For both simulation and real-robot experiments, we work with images of size 480x480. We evaluated the effectiveness of our framework by assigning our robots three distinct rearrangement tasks in cluttered environments. These tasks included shuffling objects, sorting objects (including arranging objects in a specific pattern) and random placement of objects. Visual examples of these tasks can be seen in Figure \ref{fig:tasks}. Additionally, we introduce domain randomization in different aspects of the environment to assess the robustness of our framework. Some of these variations are outlined below:
\begin{itemize}[leftmargin=*]
\item Size of the arena: 4.5 meters to 5.5 meters.
\item Number of agents: 2 and 3.
\item Number of objects: 8, 12, and 16.
\item Object colors: Red, green, and blue, with Gaussian noise.
\item Number of static obstacles: Ranging from 5 to 14.
\item Arena color: Gray values ranging from 128 to 255.
\end{itemize}
The size of agents was fixed at 0.4 meters, while the size of objects was kept fixed at 0.1 meters.

\subsection{Data Generation: Training and Testing}
We utilized PyBullet module to gather our dataset, comprising 600 distinct environments. Within each environment, we define 20 different configurations, distributed as follows: 40\% for shuffling, 30\% for sorting, and 30\% for random object placement tasks. To obtain accurate ground truth data for our training dataset, we employed specific methodologies.  For the object selection module, we utilized the A* algorithm on each agent to compute the shortest path from its location to every object in the environment, detected using instance segmentation. This process generated the ground truth heatmap, which served as a reference for object selection. In order to account for objects that were already positioned at their target locations, we multiplied those locations with a factor of 0.1 in the ground truth heatmap. This adjustment reflected the lower likelihood of these objects being moved in most cases. For the feasibility module, we sampled a set of locations within the workspace. Applying the A* algorithm from potential object pickup locations to the sampled locations allowed us to generate a heatmap. Areas with high density were scaled with a factor of 0.5, considering the lower feasibility for subsequent rearrangement actions. In the quality network module, we adopted a heuristic approach using the Euclidean distance from the closest target location of the pickup object to every location in the workspace. This calculation enabled us to compute the ground truth heatmap, which served as the reference label for the quality network. To increase the size of our dataset and improve network performance, we also applied data augmentation techniques such as vertical and horizontal flips, as well as clockwise and anticlockwise rotations by 90 degrees. From all data samples, we use 80\% for training and the remainder for evaluation. Also note that we train our models in the simulation with domain randomization and use our real-world setup only for evaluation purposes to demonstrate the sim-to-real transfer capability of our framework.

\subsection{Training Details}
\subsubsection{Model Architectures}
The implementation of our algorithm was carried out using PyTorch. One of the key characteristic of our algorithm lies in the uniformity of the network architectures employed. Each network leverages a transformer network, following the framework outlined in \cite{vaswani2017attention}. The foundation of each network is a Fully Convolutional Network (FCN) responsible for encoding the environment input into a latent space. The FCN achieves dimensionality reduction through a sequence of convolutional, ReLU, and MaxPool layers. By applying a sliding window of size $p \times p \times i$ to the input, where $i$ varies depending on the network type (7 for object selection, 2 for feasibility, and 9 for quality), the FCN generates an output of dimensions $H_l \times W_l \times d$. Here, $H_l$ and $W_l$ correspond to the height and width of heatmap, while $d$ represents the latent dimension of the transformer encoder. The value of $p$ is determined by the smallest block that an agent or object can occupy in the image. We set the patch size $p$ to 20 for our experiments. The output of the FCN is subsequently passed through the transformer encoder, which facilitates the learning of connections between local regions in the current and target scenes of the environment. The encoder architecture encompasses a series of multi-headed self-attention (MSA) and multi-layer perceptron (MLP) blocks. To further enhance the model's performance, we employ Dropout, Layer Normalization, and residual connections, following the approaches presented in \cite{vaswani2017attention} and \cite{dosovitskiy2020image}. Through its attention mechanism applied to all patches, the network effectively encodes the significance of individual pixels for the specific task at hand. Lastly, the output of the encoder undergoes a $1 \times 1$ convolutional layer, enabling the prediction of probabilities for each patch.

\subsubsection{Parameters}
We use the L2 loss and the Adam optimizer with $\beta_1 = 0.9$, $\beta_2 = 0.98$, and $\epsilon = 1e-9$. We varied the learning rate as proposed in \cite{vaswani2017attention} with warm-up steps of 3200. The latent size $d$ of the transformer encoder was set to 512. We apply residual dropout of 0.1 as presented in \cite{vaswani2017attention}. Each model was trained for 100 epochs with a batch size of 120. The models were trained on one machine with 3 NVIDIA 3090RTX graphics card.

\section{Results}
Our evaluation consisted of four sets of experiments. Firstly, we evaluated the performance of our method on various rearrangement scenarios, comparing it against several classical baselines. The number of agents and objects used in these scenarios was consistent with the training data. Secondly, we show the generalization capability of our method by testing it on object rearrangement tasks with unseen numbers of objects and agents. Thirdly, we conducted an ablation study to analyze the individual contributions of each component within the MANER framework. Lastly, we showcased the sim-to-real generalization of our method by applying it to real-world object rearrangement tasks. To quantitatively compare the performance of different methods, we employed the following metrics:

\begin{itemize}[leftmargin=*]
\item \textbf{Placement Success Rate (SR)}: This metric represents the percentage of successfully placed objects at their desired target positions.
\item \textbf{Distance Traveled (DT)}: This metric measures the total distance covered by all agents in rearranging the objects from the source configuration to the target configuration.
\item \textbf{Completion Time (CT)}: This metric quantifies the time spent by the agents to complete the object rearrangement process from the source configuration to the target configuration.
\item \textbf{Inference Time (IT)}: This metric indicates the total time spent by the model in planning all the rearrangement actions using 1 CPU. The maximum allowable planning time is set to 120 seconds.
\end{itemize}

Note that the placement success rate is computed across all rearrangement scenarios, while the  distance travelled, completion time and inference time are computed only on successful rearrangements.

\subsection{Algorithm Comparison}
To evaluate the effectiveness of our approach, we conducted experiments on a range of simulated object rearrangement scenarios. These experiments involved different numbers of objects and agents, specifically 8, 12, and 16 objects with 2 and 3 agents. For each combination of agent and object, we performed 50 experiments encompassing various tasks, as depicted in Figure \ref{fig:tasks}. Tables \ref{tab:sr-results} and \ref{tab:other-metrics} shows how our method compares to multiple baseline methods. The baselines we evaluated include:

\begin{table}[ht]
\caption{\textbf{Placement Success Rate.} We show the success rate (\%) averaged over 2 and 3 agents for different number of objects (8, 12, 16). Higher is better.}
\label{tab:sr-results}
\centering
\begin{tabular}{cccc}
  \toprule
  \multirow{2}{*}{Algorithms} 
      & \multicolumn{3}{c}{\# Objects}     \\ \cmidrule{2-4}
  & 8 & 12 & 16 \\  \midrule
 Classical (Random) & 76.25 & 65.33 & 48.12 \\      
 Classical (Greedy) & 77.63 & 66.41 & 53.06 \\   
 MANER (Ours) & \textbf{81.25} & \textbf{76.33} & \textbf{75.93} \\ \bottomrule
\end{tabular}
\vspace{-0.35cm}
\end{table}

\begin{table}[ht]
\caption{\textbf{Average Distance Traveled (meters) and Completion Time (seconds).} We show the results for different sets of object-agent number pair. Lower is better.}
\label{tab:other-metrics}
\centering
\begin{tabular}{ccc|c|c|c}
  \toprule
  \multirow{3}{*}{\# Objects} & \multirow{3}{*}{Algorithms} 
      & \multicolumn{2}{c|}{2 Agents} & \multicolumn{2}{c}{3 Agents} \\ \cmidrule{3-6} &

  & DT & CT & DT & CT \\ \midrule
 \multirow{2}{*}{8} & Classical (Random) & 41.1 & 144.7 & 38.4 & 97.7 \\
 & Classical (Greedy) & 35.2 & 117.4 & 32.1 & 85.3 \\ 
 & MANER (Ours) & \textbf{24.6} & \textbf{82.3} & \textbf{24.4} & \textbf{57.4} \\
 \midrule
 \multirow{2}{*}{12} & Classical (Random) & 64.5 & 218.4 & 68.2 & 166.5 \\     
 & Classical (Greedy) & 51.0 & 176.8 & 59.6 & 160.5\\ 
 & MANER (Ours) & \textbf{36.6} & \textbf{116.4} & \textbf{37.5} & \textbf{87.5} \\ 
 \midrule
 \multirow{2}{*}{16} & Classical (Random) & 77.2 & 265.8 & 59.9 & 144.7\\     
 & Classical (Greedy) & 65.1 & 
 231.6 & 50.4 & 123.0  \\ 
 & MANER (Ours) & \textbf{48.1} & \textbf{159.8} & \textbf{39.0} & \textbf{98.1} \\
 \bottomrule
\end{tabular}
\vspace{-0.2cm}
\end{table}

\begin{figure}[ht]
\centering
\includegraphics[trim={0.0cm, 0.0cm, 0, 0.0cm}, clip, width=0.90\columnwidth]{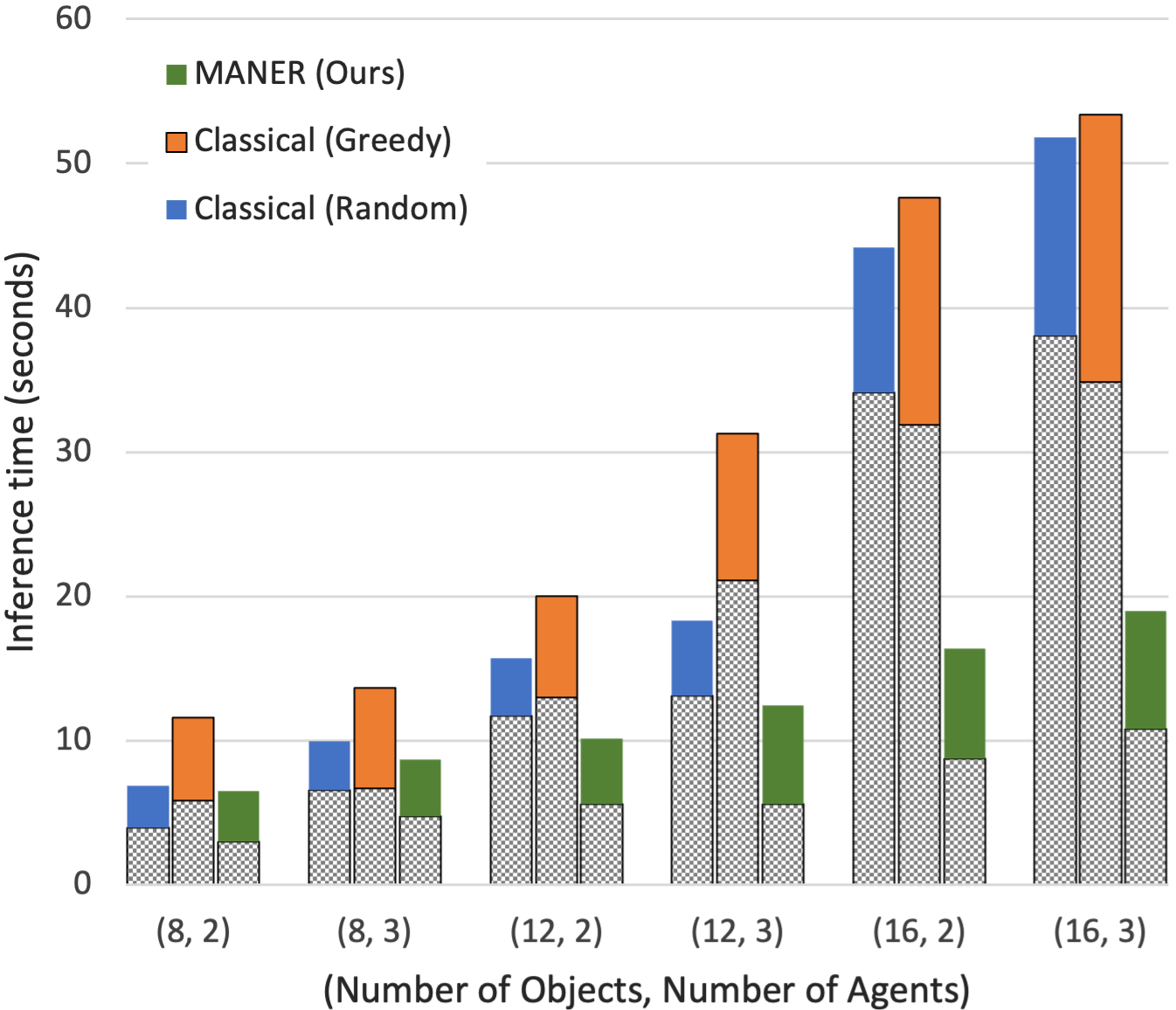}
\vspace{-0.0in}
\caption{Inference Time Comparison of Baseline Models and MANER on evaluated dataset with varying numbers of objects and agents. The time taken by path planning component is indicated as gray shaded area. The inference time for MANER increases almost linearly with increasing number of objects and agents, while it grows exponentially for classical models.}
\label{fig:execution-time}
\vspace{-0.3cm}
\end{figure}

\begin{figure*}[ht]\vspace{0.06in}
\centering
\includegraphics[trim={0 0 2cm 0.1cm}, clip, width=\textwidth]{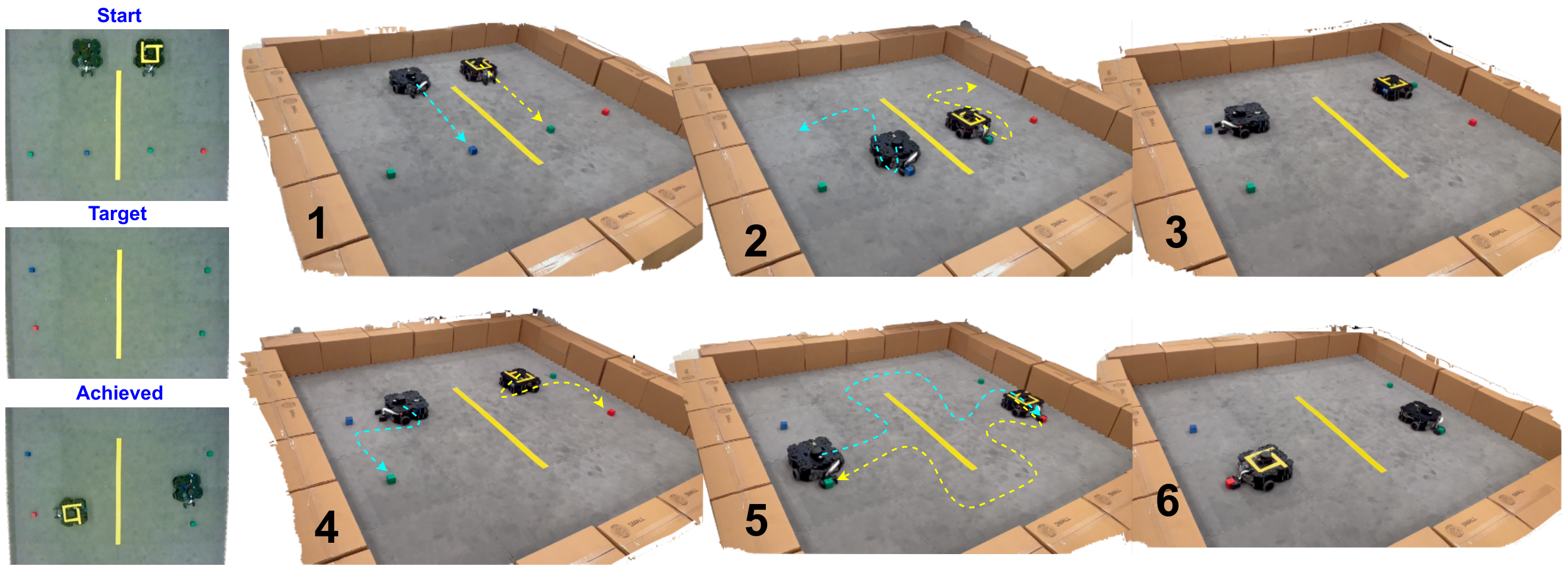}
\caption{Example of a real-world experiment with 4 objects and 2 agents. The images on the left show the start, target, and achieved configurations. The images labeled with numbers on the right show the intermediate configurations achieved by agents based on the trajectories from MANER.}
\label{fig:real-world-experiment}
\vspace{-0.0cm}
\end{figure*}

% \begin{figure*}[ht]
% \centering
% \includegraphics[width=\textwidth]{6objnew2.png}
% \caption{Example of a real-world experiment with 6 objects and 2 agents. The images on left show the start scene, target scene and final achieved scene. The images labeled with numbers on the right show the intermediate configurations achieved after agents are navigated based on the trajectories obtained using MANER.}
% \centering
% \label{fig:real-world-experiment}
% \end{figure*}

\subsubsection{Classical (random)}
This baseline method is a multi-agent variant of a search-based approach described in \cite{krontiris2015dealing} for addressing non-monotone instances of object rearrangement. The method employs instance segmentation to detect objects and agents in the start and target scenes. Each agent is assigned a specific object, and the method determines the closest unoccupied target location for each object from the target scene. If the resulting multi-agent path, computed using \cite{yakovlev2017any} is free from collisions, the method proceeds iteratively to rearrange the remaining objects. However, if the computed path is obstructed, the method attempts to clear the path by randomly sampling available regions in the environment as potential relocation spots for the obstructing objects. It then moves the objects to the sampled locations. This process continues until a valid rearrangement solution is found or the time limit is exceeded.

\subsubsection{Classical (greedy)}
This baseline method also begins with instance segmentation to identify objects and agents in the current and target scenes. The objects are then mapped one-by-one in the scenes using a bipartite graph matching algorithm based on computed A* distances between same-colored objects. Each agent is assigned an object from the start scene. The transfer locations for each agent-object pair are determined based on one-to-one mapping. If the resulting multi-agent path is free from collisions, the algorithm proceeds to rearrange the objects. If the path is obstructed, the method attempts to clear the path by sampling the closest available regions in the environment as potential relocation spots for the obstructing objects. It then moves the objects to the sampled locations. The process continues until a valid rearrangement solution is found or the time limit is exceeded.

\textbf{Comparison to classical methods}: Table \ref{tab:sr-results}, Table \ref{tab:other-metrics}, and Figure \ref{fig:execution-time} highlight the superior performance of MANER compared to the baseline methods. This performance improvement can be attributed to several factors. The classical approach often leads to collisions due to random object selection and placements, especially in cluttered environments. In contrast, MANER utilizes transformer networks and self-attention mechanisms, which effectively capture spatial relationships and dependencies among objects, agents, and static obstacles in the environment. This enables it to identify optimal object-agent pairs and corresponding regions, resulting in a higher success rate and reduced traversal time. Furthermore, the classical approach exhaustively considers all possible rearrangement actions, leading to longer execution times in the worst case. In contrast, the execution time of MANER grows linearly with the increasing number of agents, providing a more efficient solution. Also worth mentioning that most objects which MANER fails to rearrange are due to reachability constraints, as some objects cannot be picked or dropped since objects, agents, and static obstacles are positioned randomly as discussed in \ref{section:env-setup}.

\textbf{Generalization to Different Number of Objects and Agents}: Table \ref{tab:generalization} demonstrates MANER's generalization capability across different scenarios involving varying numbers of objects and agents. While the training dataset for MANER comprised scenes with 2 and 3 agents and 8 to 16 objects, the evaluation results show that MANER performs effectively in challenging environments with 4 agents and a broader range of 8 to 24 objects. The success rate remains relatively consistent even with the addition of more agents to the scene, and only exhibits a slight decrease as more objects are introduced. Additionally, the total distance traveled by the agents remains similar, while the completion time decreases linearly, indicating that the rearrangement task is evenly distributed among all agents. These findings highlight MANER's robustness in handling scene clutter and efficiently solving complex rearrangement tasks.

\begin{table}[ht]
\caption{Generalization to unseen number of objects and agents.}
\label{tab:generalization}
\centering
\begin{tabular}{cccc}
  \toprule
  \multirow{2}{*}{\#}
      & \multicolumn{3}{c}{Performance Metrics} \\ \cmidrule{2-4}
  & SR (\%) & DT (m) & CT (sec) \\ \midrule
 8 objects, 4 agents & 82.5 & 25.9 & 52.6 \\   
 12 objects, 4 agents & 77.16 & 37.4 & 73.8 \\ 
 16 objects, 4 agents & 73.5 & 48.5 & 107.6 \\ 
 24 objects, 4 agents & 68.83 & 65.2 & 154.6 \\ 
 \bottomrule
\end{tabular}
\vspace{-0.4cm}
\end{table}

\subsection{Ablation Studies}
We conducted a series of experiments using 12 objects and 3 agents to analyze the impact of different components in our method. Specifically, we examined the impact of $k$, the object selection network, and the region proposal network. The results are presented in Table \ref{tab:ablation}. The findings indicate the significance of all these components. When only the best region from the region proposal network is selected, the placement success rate drops slightly. When object selection is absent, the model randomly chooses objects for drop-off, resulting in a reduced success rate and increased distance traveled. Without the region proposal, the success rate decreases drastically as random regions are proposed instead of carefully selected regions for object transfers.

\begin{table}[ht]
\caption{Ablation analysis of the various components of the network}
\label{tab:ablation}
\centering
\begin{tabular}{cccc}
  \toprule
  \multirow{2}{*}{Algorithms} 
      & \multicolumn{3}{c}{Performance Metrics} \\ \cmidrule{2-4}
  & SR (\%) & DT (m) & CT (sec) \\  \midrule
 MANER & 79.9 & 35.2 & 89.9 \\
 w/ Region Proposal ($k=1$) & 72.5 & 40.12 & 99.1 \\ 
 w/o Object Selection & 69.5 & 38.4 & 96.6 \\   
 w/o Region Proposal & 28.87 & 58.5 & 168.1 \\ 
 \bottomrule
\end{tabular}
\vspace{-0.25cm}
\end{table}

\subsection{Real Robot Experiments}
Finally, we conducted a series of real-world experiments using two TurtleBot3 robots in an arena with a size of 2.4 meters. Due to size constraints on the robots and the environment, the experiments were limited to scenarios with 4 and 6 objects. As depicted in Fig. \ref{fig:real-world-experiment}, the robots were tasked with rearranging a scene consisting of 6 objects to achieve a desired target configuration. The real-world tasks presented additional challenges, including noisy scene segmentation and odometry drift caused by factors such as wheel slippage and sensor noise.  These challenges occasionally resulted in rearrangement failures. Nonetheless, MANER demonstrated remarkable generalization to real-world settings with IT $9.5 \pm 1.6$ seconds, DT $19.6 \pm 5.4$ meters, and CT $180 \pm 32$ seconds across various rearrangement setups. Additional examples can be found in the supplementary materials.

\section{Conclusion, Limitations, and Future Work}
We presented the Multi-Agent Neural Rearrangement Planning (MANER) approach that rearranges objects to their desired configurations in cluttered environments. We evaluated MANER on challenging problems and demonstrated its sim-to-real generalizations. One challenge to our approach is the precise low-level control since pixel-to-world coordinate mapping can result in precision errors and collisions, leading to failures. Another challenge is from our path planning module, which currently does not tackle orientation constraints during object placements. Hence, for future works, we envision building precise motion planning and control strategies to solve even harder multi-agent object rearrangement tasks.
%We presented Multi-agent Neural Rearrangement Planning (MANER), a learning-based object rearrangement approach for cluttered environments. MANER works in an end-to-end fashion and can rearrange a configuration to a desired configuration given the start and target scenes of the environment. We evaluate MANER on challenging problems and demonstrate its sim-to-real generalizations. {\color{red} One challenge to our approach is the precise control since pixel-to-world coordinate mapping can result in precision errors and collisions leading to rearrangement failures. Hence, for future work, we envision having a fine-grained pixel coordinate system for precise low-level control to solve even harder rearrangment tasks.% We also want to scale our approach to handle dynamic environments with arbitrary objects and explore its application to warehouse rearrangement tasks.}

% \begin{table}[h!]
% \centering
% \begin{tabular}{|p{3cm}|p{1.5cm}|p{1cm}|p{1cm}|p{1cm}|}
%  \hline
% Number of agents/ Number of objects & 4 & 8 & 12 & 16 \\ [0.5ex] 
%  \hline\hline
% 1 & 6 & 87837 & 787 & 23 \\
%  \hline
% 2 & 7 & 78 & 5415 & 23 \\
%  \hline
% 4 & 545 & 778 & 7507 & 23 \\
% \hline
% 8 & 545 & 18744 & 7560 & 23 \\[1ex]
% \hline
% \end{tabular}
% \caption{Time to complete target formation for our pipeline with varying number of agents and varying number of objects.}
% \label{table:1}
% \end{table}

\bibliographystyle{IEEEtran}
\bibliography{IEEEfull}


%%%%%%%%%%%%%%%%%%%%%%%%%%%%%%%%%%%%%%%%%%%%%%%%%%%%%%%%%%%%%%%%%%%%%%%%%%%%%%%%

%%%%%%%%%%%%%%%%%%%%%%%%%%%%%%%%%%%%%%%%%%%%%%%%%%%%%%%%%%%%%%%%%%%%%%%%%%%%%%%%



% \section*{ACKNOWLEDGMENT}

% The preferred spelling of the word ÒacknowledgmentÓ in America is without an ÒeÓ after the ÒgÓ. Avoid the stilted expression, ÒOne of us (R. B. G.) thanks . . .Ó  Instead, try ÒR. B. G. thanksÓ. Put sponsor acknowledgments in the unnumbered footnote on the first page.

% %%%%%%%%%%%%%%%%%%%%%%%%%%%%%%%%%%%%%%%%%%%%%%%%%%%%%%%%%%%%%%%%%%%%%%%%%%%%%%%%

% References are important to the reader; therefore, each citation must be complete and correct. If at all possible, references should be commonly available publications.

\end{document}